# Multi-Agentic System Leveraging Open-Source LLMs to Mitigate Disinformation Threats

Sebastian Kula[1,2] and Martin Tamajka[1]

[1] Kempelen Institute of Intelligent Technologies, Bratislava, Slovakia
{sebastian.kula,martin.tamajka}@kinit.sk
[2] West Pomeranian University of Technology in Szczecin, Szczecin, Poland

**Abstract.** In contemporary societies, the threat of disinformation has reached alarming levels, exacerbated by the proliferation of electronic communication, social media, and advancements in artificial intelligence. As a result, there is an urgent need to develop effective countermeasures to mitigate this menace. However, the sheer scale of the problem renders manual fact-checking and human-based verification inadequate, underscoring the necessity for automated methods to detect and debunk disinformation. This article proposes a novel approach based on a multi-agent system that emulates the decision-making processes of human annotators engaged in disinformation detection tasks. By incorporating a consensus mechanism, diversity in cognition and diversity in knowledge, and also hierarchical structure, inspired by human annotators' behavior, the proposed method achieves superior results compared to individual Large Language Models (LLMs), including GPT 4 and GPT 3.5. The system leverages open models (e.g., LLaMA, Kimi, Qwen, Deepseek and LLaMA-Nemotron) to ensure greater transparency. The evaluation of the proposed method encompasses datasets in languages with varying resource availability, including English (high-resource), Polish (medium-resource), Slovak (low-resource) and Bulgarian (low-resource). Experiments were conducted on tasks such as direct disinformation detection, identification of texts worthy of verification, and detection of texts containing verifiable factual claims.

**Keywords:** LLMs · disinformation detection · multi-agent system · open models · NLP· automated data annotation.

## 1 Introduction

The spread of disinformation has become a pressing concern in today's digital age, with far-reaching consequences for individuals, communities, and societies as a whole. The proliferation of social media platforms, coupled with the increasing sophistication of artificial intelligence, has created an environment in which false information can spread rapidly and with relative ease. To combat this threat, it is essential to develop innovative and effective methods for detecting and debunking disinformation. Traditional approaches relying on human resources for manual verification are no longer sufficient due to the sheer scale and velocity of information dissemination. Consequently, there is an ongoing and urgent need to develop automated methods capable of detecting and exposing disinformation effectively.

This article proposes a novel methodology based on a multi-agent system (MAS) designed to emulate cognitive decision-making processes typically employed by human annotators in disinformation mitigation tasks. Specifically, the model incorporates consensus-building mechanisms derived from observations of how human experts collaboratively assess and validate potentially misleading content. Additionally, hierarchical reasoning patterns observed among human annotators have been integrated into the multi-agent framework to reflect real-world decision-making structures. The design of the proposed system incorporated also consideration of variability in annotators' knowledge and cognitive abilities.

The proposed approach demonstrates superior performance compared to standalone LLMs, including commonly used models such as GPT-4 and GPT-3.5 [25]. Importantly, the multi-agents leverage open-source LLMs - including LLaMa [13], Kimi [30], Deepseek [9], Qwen [32], and LLaMa-Nemotron [4], which enhances transparency, reproducibility, and ethical accountability within the system's design. The use of multilingual models enables cross-linguistic applicability, and the method was evaluated across datasets representing languages with varying levels of availability of linguistic resources: high-resource (English), mid-resource (Polish), and low-resource (Slovak and Bulgarian). The evaluation of the proposed method across languages with varying resource availability highlights its potential for widespread applicability. The results demonstrate the potential of the system as a valuable tool in the fight against disinformation. Furthermore, the system's ability to detect check-worthy texts and texts containing verifiable factual claims offers a promising path for further research and development.

In order to systematically assess and visualize the impact of the proposed methods, the following research questions have been stated:

*RQ1* To what extent can a MAS effectively replicate the performance of human annotators in highly specialized disinformation detection tasks when applied to small data and limited datasets? *RQ2* In the context of small data disinformation detection, can MAS built upon open models achieve comparable performance to those built upon standalone closed (proprietary) models? *RQ3* Can a MAS accurately predict disinformation characteristics in unseen, small datasets? *RQ4* Is it possible to effectively model mechanism of consensus among human annotators within debates using a MAS? *RQ5* Is it possible to effectively model and analyze the phenomenon of human annotators knowledge diversity within debates using a MAS framework?

To address these questions, a detailed analysis was conducted examining MAS and standalone LLMs as applied to tasks focused on disinformation mitigation.

This work presents several novel contributions to the field of disinformation and fake news detection. These include:

- **A Multi-Agent System for Comprehensive Disinformation Analysis:** We developed, implemented and evaluated the MAS capable of addressing a diverse range of tasks related to combating disinformation.
- **Highly Generalizable System Architecture:** We propose a system demonstrating significant generalization capabilities across linguistic resources. The proposed approach exhibits robust performance in high-, mid-, and low-resource languages.
- **Consensus-Driven Decision Making Inspired by Human Annotation:** The system's design is fundamentally grounded in the principle of consensus, mirroring human annotation processes where collective agreement strengthens decision accuracy.
- **Heterogeneous LLM Integration for Enhanced Knowledge Representation:** Unlike typical multi-agent systems employing a uniform LLM across all agents, the proposed architecture utilizes a heterogeneous configuration based on diverse open-source models.
- **Transparent Debate Logging for Enhanced Interpretability and Novel Reporting:** We provide debate transcript generated during the multi-agent deliberation process, enabling detailed observation and analysis by human annotators.

## 2 Related Work

Disinformation detection, the mitigation of false information, and the analysis of tools designed to counter fake news dissemination represent a burgeoning area of research. Current methodologies can be broadly categorized as follows: feature engineering approaches employing established machine learning algorithms such as Logistic Regression, Decision Trees, Gradient Boosting, Random Forest [15][28], Convolutional Neural Networks (CNN) and Recurrent Neural Networks (RNN) [8], Naive Bayes, Support Vector Machines (SVM), and K-Nearest Neighbors (KNN) [29] for feature extraction from textual or multimodal data to train disinformation detection models; with the development and emergence of transformers the application of transformer architectures fine-tuned for classification and detection tasks related to disinformation [17]; direct utilization of Large Language Models (LLMs) functioning as independent disinformation classifiers [26]; and agent-based systems leveraging LLMs and supplementary tools to enhance disinformation detection processes [7][19][5]. Furthermore, research is increasingly focused on multi-agent systems that employ collaborative reasoning via multiple LLM-based agents to detect disinformation content [3][20][12][18]. Among these approaches, multi-agent methodologies represent a nascent field, evidenced by the limited existing literature. However, they demonstrate significant potential for automation and minimization - or even elimination - of human intervention in disinformation detection tasks [20].

Current research within the domain of automated disinformation detection and mitigation demonstrates an increasing trend toward sophisticated system architectures. In [7] agent for misinformation detection was applied, the LLM agent leveraging a multi-tool approach. Specifically, it integrates three key tools: a precise web search tool, a source credibility assessment tool, and a numerical claim verification tool. This allows for dynamic interaction with diverse web sources, evaluation of information provenance, and robust fact-checking capabilities. The agent was implemented using LangGraph framework and GPT-4o, testing were conducted on English data [7]. The system presented in [19] introduces FactAgent, the agentic approach leveraging LLMs for fake news detection without requiring training data. Unlike prior LLM applications employing direct prompting, the FactAgent emulates human fact-checking expertise through a structured workflow. The FactAgent integrates both LLM internal knowledge and external resources (URL) for comprehensive claim verification. The FactAgent was implemented with the LangChain framework and gpt-3.5-turbo model and tests were conducted on English datasets [19].

The MCP-Orchestrated Multi-Agent System for Automated Disinformation Detection presented in [3] demonstrates significant advancements in combating information manipulation through a novel ensemble approach. The study uniquely leverages a MAS orchestrated by the Model Context Protocol (MCP) to achieve 0.953 accuracy and an F1 score of 0.964, surpassing both individual agent performance and traditional methods. The core novelty lies in the synergistic integration of four distinct agents: a machine learning agent employing logistic regression, a Wikipedia knowledge check agent utilizing named entity recognition, a coherence detection agent powered by LLM prompt engineering, and a web-scraped data analyzer extracting relational triplets for fact-checking. The system was based on LLaMa3 8B models and tested on English.

In [20] TruEDebate (TED) is presented. TED is MAS leveraging LLMs to enhance both interpretability and effectiveness in fake news detection, addressing limitations of traditional classification models and single-prompt LLM approaches. TED uniquely simulates a rigorous debate process inspired by formal debate settings. Experimental validation was conducted on English and Chinese datasets. The TED utilized GPT-4o-mini as the LLM for its agents.

Another MAS dedicated to claim verification is presented in [12]. The core novelty of the MAS lies in its two-layered framework - a Reasoning Layer comprised of specialized LLM agents executing distinct tasks, and a Decision Layer synthesizing their responses to determine claim veracity. The system demonstrably outperforms baseline models like BERT-Base, XLNet, and even single LLMs (GPT2-XL), achieving state-of-the-art performance on the FEVER dataset with 0.853 accuracy and 0.853 F1-Macro score. The agents in the system are based on the open-source LLM Gemma2-9B-IT. For the evaluation purposes the FEVER dataset was adopted.

The paper [18] introduces a multi-agent debate system where multimodal agents collaboratively assess contextual consistency and leverage external information retrieval for enhanced reasoning. A key novelty lies in the framework's ability to achieve state-of-the-art accuracy without requiring expensive finetuning, demonstrating robust performance even with minimal prior knowledge. Authors utilized preliminary open-source LLaVA model to create agents and in further experiments replaced LLaVA with GPT-4o (OpenAI). In their experiments they have conducted tests on multimodal English based datasets.

## 3 Proposed open-source multi-agent system

The proposed MAS aims to emulate the cognitive processes employed by human annotators in disinformation detection and labelling tasks. Inspired by studies detailing human annotation workflows [2] [14], this system replicates the iterative stages and decision-making methodologies observed in these processes. A core element of these workflows is consensus building through debate: final decisions are ideally reached via collective conviction based on exchanged knowledge and reasoned justifications. When full consensus proves unattainable, a mechanism for adjudication - often involving an additional annotator - is employed to arrive at a justified decision following deliberation. Recognizing that human annotation inherently involves individuals with varying levels of expertise and perspectives, the proposed method leverages MAS (with different LLMs) to simulate this diversity. To ensure that the diversity is caused by the LLM itself, all the annotation agents are instructed using the same prompt. Agents within the system generate decisions accompanied by supporting justifications, which are accessible both to a designated leader (who also moderates discussions) and to all other agents. In instances of divergent opinions, the moderator facilitates continued dialogue with the goal of achieving consensus across the agent population. To further reflect the heterogeneity of human annotators, the agents are instantiated using LLMs differing in their architectural characteristics and pre-training data. This approach acknowledges that LLMs, by virtue of their training datasets and parameterization, inherently possess diverse knowledge bases and capabilities – mirroring the phenomenon observed in real-world annotation teams. The diversity inherent within these models is well documented. For instance, the LLaMA 3 pre-training data exhibits a significant bias towards English tokens than non-English tokens [13], while DeepSeek-V3 was trained on a multilingual corpus with a strong emphasis on English and Chinese [9]. This diversity manifests in performance variations across different tasks. Specifically, LLaMa-3.1-Nemotron-Ultra-253B-v1 demonstrates superior performance on Scientific Reasoning tasks compared to the LLaMa-3.1-405B model, which excels instead at Instruction Following [4].

Figure 1 details the general conceptual framework of this work. While related approaches have been explored in prior research [18] [20], the method presented here distinguishes itself through several key innovations. Primarily, consensus-building is central to our design. It represents the primary objective, with debate serving as a facilitative mechanism rather than an end in itself. The core novelty lies in the utilization of heterogeneous LLMs for agent instantiation. This approach ensures that justifications generated by agents are grounded in diverse knowledge bases, reflect varying model capabilities relevant to specific tasks, and employ distinct reasoning methodologies. In contrast, many existing MASs rely on a uniform LLM across all agents [18] [20]. A further distinguishing feature of this work is its adaptation and evaluation on datasets encompassing not only English but also Slovak, Bulgarian and Polish. The method is specifically designed for tasks including direct disinformation detection, verifiable factual claim detection and check-worthy claim detection. Moreover, the agents are based on open-source LLMs, facilitating customization, and potentially reducing operating costs. The proposed system operates in a closed configuration, deliberately excluding any agent capable of retrieving external data from the internet or other data sources. This design choice is motivated by the predictive nature of the disinformation detection task. The system aims to identify disinformation before it has been debunked by professional fact-checkers or fact-checking portals, necessitating reliance on internal knowledge and reasoning capabilities.

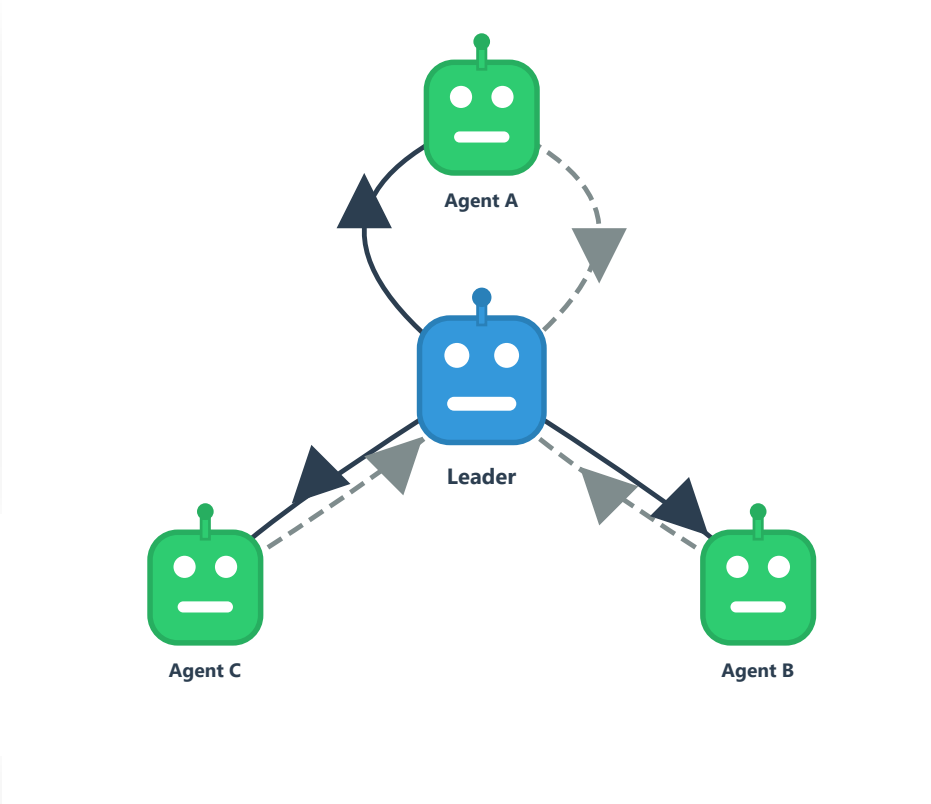

**Fig. 1.** The architecture of the proposed MAS with leader agent in the center. The architecture represents the star (hub-and-spoke) topology with agents based on various LLMs.

The proposed system comprises four agents: a designated leader agent and three expert annotator agents operating as equals. The leader assumes responsibility for moderating discussions within a hierarchical structure, functioning as the central coordinating entity. Task assignment, question formulation, and communication with the annotator agents are exclusively managed by the leader. Furthermore, the leader initiates and concludes debates and is responsible for structuring the final outcome in JSON format. This architecture establishes a star (hub-and-spoke) topology, with the leader serving as the central node, as illustrated in Figure 1. The three annotator agents function as subject matter experts, responding to prompts and tasks assigned by the leader. They provide answers to posed questions, accompanied by justifications integrated within their responses, and may be specifically requested by the leader to justify their reasoning. Communication between the annotator agents (A, B, C) is indirect, mediated solely through the leader; any reference to or consideration of another agent's statement or justification occurs only when explicitly indicated and instructed by the leader via a task assignment. The system operates under three possible scenarios, illustrated in Figure 2. In the first scenario, the leader poses a question and initiates content annotation. Annotators respond directly to the leader's query; when responses are identical - indicating full consensus - the leader immediately structures the result into JSON format, concluding the round. The second scenario is triggered when initial responses reveal a lack of consensus. The leader then initiates a debate intended to resolve these discrepancies. This iterative process continues until consensus is achieved, evidenced by agents modifying their initial annotations and justifications. A final result is then issued, terminating the debate. The third scenario addresses situations where debate fails to yield consensus. In such cases, the leader interrupts the discussion and, based on the content and justifications provided by the annotator agents, independently draws a conclusion and records the final outcome. The leader's primary function is to facilitate consensus-building through strategic shaping and moderation of the debate, aiming for rapid resolution. However, in instances where consensus remains unattainable, the debate is prolonged up to a maximum of 20 rounds; empirical observations suggest this limit provides sufficient iterations for either achieving consensus or definitively determining its unattainability. The leader's prompt incorporates explicit authorization to make a final decision when consensus cannot be reached. The annotator agents' prompts always include instructions to generate justifications for their decisions. If the leader deems it appropriate, they can ask individual agents to comment on the other agents' justifications. At the same time, the annotator agents can see the entire discussion and, as a result, can also change their decision. This means that a debate is taking place between these agents, but it is organized hierarchically and centrally by the leader. The entire debate transcript is displayed and archived, allowing for observation by human annotators. However, the system does not anticipate direct intervention by human annotators in the discussion, but these capabilities allow it to be used as direct support for human annotators.

## 4 Experiments and results

To rigorously evaluate the proposed MAS, a comprehensive experimental framework was established utilizing four benchmark datasets representing diverse facets of disinformation and fact verification. This evaluation aimed to assess performance across varying linguistic contexts (Polish, English, Slovak and Bulgarian) and data sources.

The primary dataset employed was MIPD [22], a Polish corpus specifically annotated by expert analysts in disinformation demystification, ensuring high-quality ground truth labels for reliable evaluation. Complementing this, the CLEF23 dataset [1] provided a valuable resource focused on assessing the check worthiness of textual information. Additionally, a self-collected dataset comprising claims from fact-checking portals (AFP Fakty [11] and Demagog.sk [10]) was incorporated to broaden the scope of evaluation with real-world data. Finally, the CLEF22 Twitter collection [23], labelled for verifiable claim presence, offered a social media context for assessing factual claim detection capabilities.

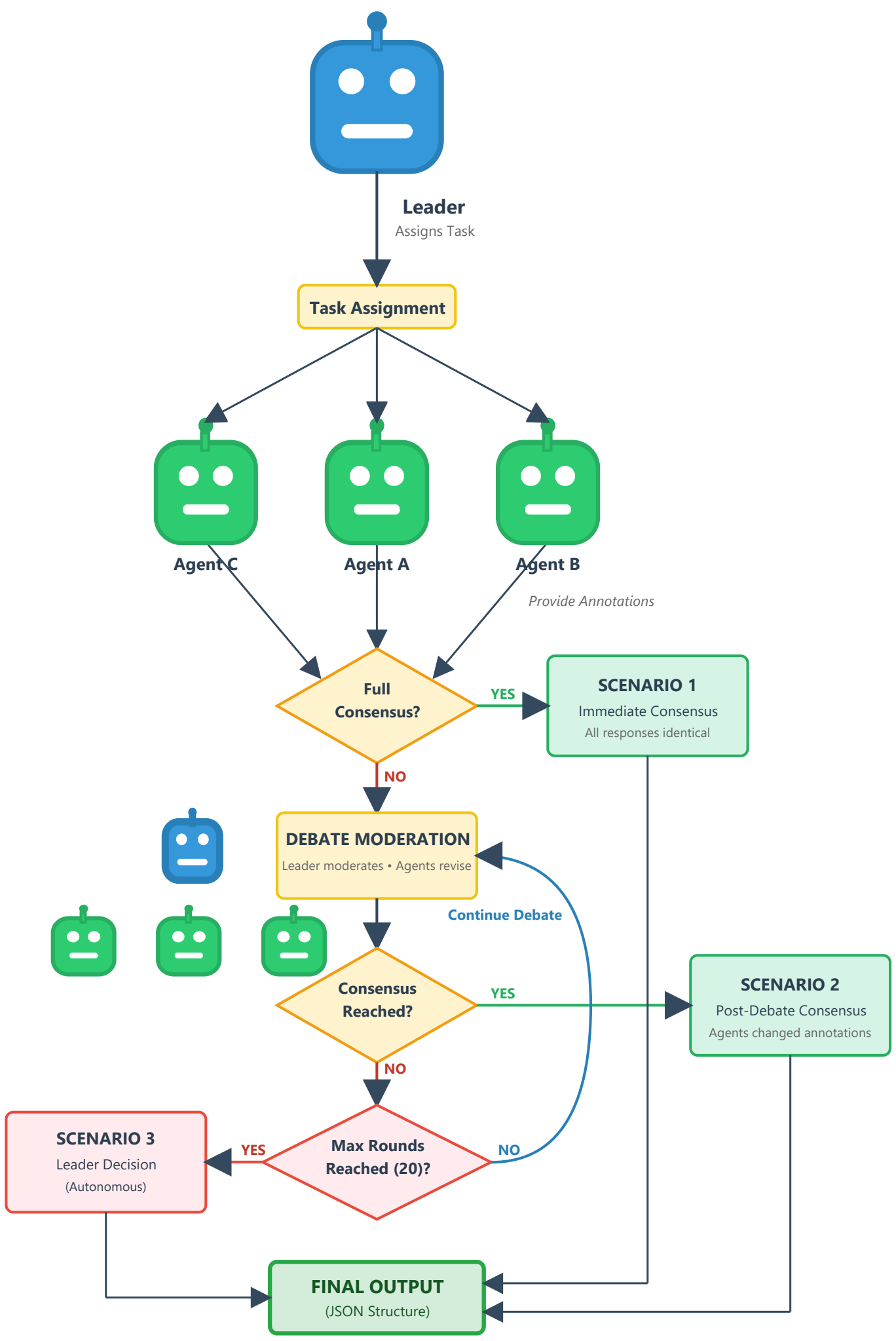

**Fig. 2.** Decision and data flow with possible scenarios in the proposed system.

Four distinct experimental tasks were designed to leverage these datasets, enabling a comprehensive assessment of the proposed system's capabilities. TASK 1 focused on disinformation detection within the MIPD dataset, employing a randomly selected 10% subset of the original testing data designated as 10pro

_disinformation_test. TASK 2 concentrated on detecting check-worthy claims utilizing the complete CT23_1B_checkworthy_english_test_gold dataset, representing the entirety of the original CLEF23 test set. Detecting check-worthy claims is the frst step in the process of detecting fake news and disinformation [2], [16]. Subsequently, TASK 3 evaluated performance on identifying check-worthy claims within the self-collected list sourced from fact-checking portals. Finally, TASK 4 was dedicated to verifiable factual claim detection and was conducted using the complete original CLEF22 test dataset. A verifiable factual claim is a sentence claiming that something is true, and this can be verified by factual verifiable information [2]. A description of the test data is provided in Table 1.

**Table 1.** Dataset overview used for the testing tasks.

| | **TASK 1** | **TASK 2** | **TASK 3** | **TASK 4** |
|---|---|---|---|---|
| No. of Samples | 152 | 318 | 30 | 329 |
| Task Name | Dis. | CwC | CwC | VFC |
| % of Original dataset | 10 % | 100 % | 100 % | 100 % |
| Language | Polish | English | Slovak | Bulgarian |
| Original Dataset | MIPD | CLEF23 | self-collected | CLEF22 |

*Note:* Dis. (Disinformation), CwC (Check-worthy Claim), VFC (Verifiable Factual Claim).

All tasks were performed utilizing code executed within the Google Colaboratory platform [6]. The LLMs integral to the MAS were requested remotely using hosted on the NIM NVIDIA Cloud Platform models [24] and AutoGEN framework [31]. The datasets employed in these experiments are publicly accessible and available for the research purposes usage.

For the experiments four prompt groups have been created for each task; each group contains a prompt for the leader and an identical prompt for the annotator agents. The prompts are presented in the appendix in additional materials.

The experimental evaluation of the proposed MAS for direct disinformation detection in Polish (TASK 1) demonstrates strong performance across key classification metrics. The model configuration employed a hierarchical structure with a leader agent utilizing Meta's LLaMa-3.1-405B-Instruct and three subordinate agents powered by diverse open-source language models: Kimi-K2, LLaMa-3.1-405B-Instruct, and LLaMa-4-Scout [21].

On the 10pro_disinformation_test dataset, the system achieved an overall accuracy of 0.88, with a macro-averaged F1-score of 0.87 and a weighted average F1-score of 0.88. These results indicate robust generalization capabilities across both classes-disinformation (class 1) and non-disinformation (class 0). Results are presented in Table 2.

Notably, the system exhibits high sensitivity towards detecting actual disinformation instances, with a recall of 0.98 for class 1, indicating that nearly all true disinformative samples were correctly identified. This high level of recall provides robust evidence supporting *RQ1*, namely that a multi-agentic system can achieve performance comparable to human annotators in the disinformation detection task. However, precision for this class stands at 0.74, suggesting a moderate rate of false positives-instances incorrectly flagged as disinformation. In contrast, the system shows excellent precision (0.99) for non-disinformation content but lower recall (0.83), implying that some portion of legitimate texts were misclassified as disinformative.

When benchmarked against commonly used LLMs such as GPT-5.2, GPT-4 and GPT-3.5 under zero-shot settings [22], the proposed multi-agent framework outperforms baselines presented in the literature [22] but slightly underperforms to our self-conducted experiments for standalone GPT-5.2 in terms of both accuracy and F1 scores. Specifically for accuracy the MAS achieves 0.88, compering to 0.91 for GPT-5.2 and 0.86 for GPT-4 and significantly outperforming GPT-3.5, which peaks at 0.70. F1 Score (macro) at 0.87, the system exceeds GPT-4's maximum F1 of 0.77 and GPT-3.5's peak of 0.56 but is below GPT-5.2 result of 0.90. The MAS recall for class 1 (disinformative content) with a value of 0.98 demonstrates superior result compared to GPT-5.2's top score of 0.84. These comparative results give a positive answer for the *RQ2* and underscore the effectiveness of incorporating structured collaboration among multiple specialized, open LLMs based agents over reliance on standalone closed LLMs.

**Table 2.** Disinformation detection, results of the proposed system for the TASK1.

| Class | Prec. | Rec. | F1 | Sup. |
|---|---|---|---|---|
| 0 | 0.99 | 0.83 | 0.91 | 103 |
| 1 | 0.74 | 0.98 | 0.84 | 49 |
| accuracy | | | 0.88 | 152 |
| macro avg | 0.86 | 0.91 | 0.87 | 152 |
| weighted avg | 0.91 | 0.88 | 0.88 | 152 |

To evaluate the effectiveness of the proposed MAS for check-worthy claim detection in English (TASK 2), we conducted a series of controlled experiments using the benchmark dataset CT23_1B_checkworthy_english_test_gold, which consists of 318 annotated samples labelled as either Yes (check-worthy) or No (non-check-worthy). The experimental framework was designed to compare the performance of our agentic architecture against LLMs operating in isolation.

For comparative purposes, standalone evaluations were also performed using two high-performing LLMs under standard prompting strategies. The first comparison involved LLaMa-3.1-405B Instruct with Chain-of-Thought (CoT) prompting tailored for CLEF-style tasks, while the second utilized GPT-4o Mini employing a guided-answer (GA) prompt format. All models were evaluated using identical data splits and metrics including precision, recall, F1-score, and overall accuracy. Temperature settings were fixed at 0.2 across all runs to minimize stochasticity while preserving linguistic diversity in responses.

**Table 3.** Check-worthy claim detection, results of the proposed system for the TASK2.

| Class | Prec. | Rec. | F1 | Sup. |
|---|---|---|---|---|
| No | 0.93 | 0.75 | 0.83 | 210 |
| Yes | 0.64 | 0.89 | 0.75 | 108 |
| accuracy | | | 0.80 | 318 |
| macro avg | 0.79 | 0.82 | 0.79 | 318 |
| weighted avg | 0.83 | 0.80 | 0.80 | 318 |

The MAS demonstrated strong classification performance with a precision of 0.93 and recall of 0.75 for non-check-worthy claims (Class No), resulting in an F1-score of 0.83 across 210 support instances (Table 3). For check-worthy claims (Class Yes), the system achieved a precision of 0.64 and recall of 0.89, yielding an F1-score of 0.75 across 108 support instances. The overall accuracy reached 0.80, with macro average F1-score of 0.79 and weighted average F1-score of 0.80.

These results demonstrate that the MAS significantly improves upon traditional single-model inference when detecting check-worthy content. Notably, it achieves higher recall for positive cases-indicating better sensitivity toward identifying actual claims requiring verification-while maintaining strong specificity via high precision on negative examples.

Comparing TASK 2 results with Standalone LLMs: When evaluating LLaMa-3.1-405B-Instruct with CoT prompting, the model exhibited a precision of 0.95 and recall of 0.60 for non-check-worthy claims, producing an F1-score of 0.74. For check-worthy claims, it achieved a precision of 0.55 and recall of 0.94, resulting in an F1-score of 0.69. The overall accuracy was 0.72, with a macro average F1-score of 0.72.

While this variant exhibits very high recall for check-worthy instances, its poor performance in correctly classifying non-check-worthy statements indicates a tendency toward over-triggering, which is a common issue in monolithic LLM-based classifiers lacking external validation mechanisms.

The evaluation of GPT-4o Mini with guided-answer prompting revealed a precision of 0.90 and recall of 0.76 for non-check-worthy claims, generating an F1-score of 0.83. For check-worthy claims, the model achieved a precision of 0.65 and recall of 0.84, yielding an F1-score of 0.73. The overall accuracy reached 0.79, with a macro average F1-score of 0.78.

GPT-4o Mini demonstrates competitive performance comparable to the agentic system but falls slightly short in terms of macro-average F1 score. Although both systems achieve similar accuracy, the agentic setup shows superior generalization across classes due to improved handling of edge cases through inter-agent debate and consensus-building.

The performance of the proposed MAS was also evaluated in comparison to the results reported in the CLEF23 paper [27] [1]. Although the fine-tuned models presented in [27] achieved significantly better results, it is noteworthy that the MAS obtained a relatively high macro-average recall of 0.82, which is comparable to the 0.85 recall reported in [27]. Moreover, the recall value increased to 0.89 for test examples related to check-worthy claims (Class Yes). These recall values indicate that the proposed MAS is capable of detecting check-worthy claims in texts (e.g., posts) with a performance comparable to that of the best fine-tuned models reported in [27]. This is a crucial aspect in check-worthy claims detection tasks, where the primary objective is to identify texts that are worth checking. The recall metric is therefore essential in practical and real-world applications dedicated to check-worthy claims detection tasks.

To verify the potential of the proposed MAS in solving real tasks faced by human annotators, experiments were conducted within TASK 3 on dataset consisting of examples generated after the release of LLMs applied as agents. This ensured an evaluation independent of training data, mitigating potential biases related to pre-existing LLM knowledge. The experimental setup comprised four agents, each powered by distinct LLMs: "Leader" Kimi-K2, "Agent A" Kimi-K2, "Agent B" Deepseek-v3.1, and "Agent C" LLaMa-4-Scout. All agents received prompts formulated in Slovak. The primary metric used for evaluation was recall, reflecting the system's ability to correctly identify check-worthy claims. Proposed MAS achieved a recall of 0.967. This indicates a high degree of sensitivity in identifying claims that warrant further fact-checking, and effectively validating the *RQ3*. This confirms the system's predictive capability regarding check-worthy claims, potentially disinformative, even when evaluated on previously unseen and relatively small datasets, indicating promising generalizability. This result demonstrates strong performance in capturing relevant instances, minimizing the risk of overlooking potentially problematic information. The high recall achieved by proposed MAS highlights its potential as a valuable tool for automated claim verification workflows. The use of diverse LLMs within the agentic architecture appears to contribute to robust performance, potentially mitigating biases inherent in individual models.

Addressing TASK 4, the system employs a collaborative architecture featuring a leader agent based on LLaMa-3.1-405b-instruct model and three subordinate agents utilizing various LLMs: Nvidia's LLaMa-3.1-nemotron-ultra-253b-v1, LLaMa-3.1-405b-instruct, and Qwen3-235b-a22b. The system was evaluated on the CT22_bulgarian_1B_claim_test_gold dataset, comprising 329 samples representing both verifiable and non-verifiable factual claims. The results demonstrate relatively high overall performance, achieving an accuracy of 0.8. A detailed breakdown of the classification metrics is presented in Table 4. These results indicate a balanced performance across both classes. The macro-averaged F1-score of 0.79 further confirms this consistent performance. The weighted average F1-score aligns with the accuracy, suggesting that there is no significant bias towards either class due to the distribution of the dataset.

To contextualize these findings, we compare our multi-agent system's performance against results obtained from single LLM configurations on the same dataset. Our MAS demonstrably outperforms both single LLM configurations in terms of overall accuracy and F1-score (Table 5). Specifically, the improvement

**Table 4.** Verifiable factual claim detection, results of the proposed system for the TASK4.

| Class | Prec. | Rec. | F1 | Sup. |
|---|---|---|---|---|
| No | 0.84 | 0.82 | 0.83 | 199 |
| Yes | 0.74 | 0.76 | 0.75 | 130 |
| accuracy | | | 0.80 | 329 |
| macro avg | 0.79 | 0.79 | 0.79 | 329 |
| weighted avg | 0.80 | 0.80 | 0.80 | 329 |

**Table 5.** Performance comparison of standalone LLaMA models and GPT-5.2 on the Bulgarian dataset (TASK 4).

| **Model** | **Accuracy** | **F1-Score** | | **Recall** | |
|---|---|---|---|---|---|
| | | **Class No** | **Class Yes** | **Class No** | **Class Yes** |
| LLaMa-3.1-405b-instruct | 0.78 | 0.81 | 0.73 | 0.79 | 0.75 |
| LLaMa-3.1-70b-instruct | 0.73 | 0.81 | 0.54 | 0.95 | 0.39 |
| GPT-5.2 | 0.69 | 0.67 | 0.72 | 0.51 | 0.98 |

*Note:* All models tested on 329 samples with temperature=0.2.

over LLaMa-3.1-405b-instruct model is notable, suggesting that the collaborative approach enhances performance compared to relying solely on single LLaMa-3.1-405b-instruct. The significant difference between our system and LLaMa-3.1-70b-instruct highlights the benefits of utilizing a larger model (LLaMa-3.1-405b) within the multi-agent framework. Furthermore, our system's results were validate against the literature [23], the obtained accuracy of 0.80 falls between the performance of AI Rational (0.839) and TOBB ETU (0.742), representing a strong result considering the different methodologies employed. While AI Rational currently holds the highest reported accuracy, our approach achieves competitive results without relying on task-specific data augmentation techniques.

It is also important to note that models fine-tuned on downstream tasks often exhibit excellent performance on test data and in-distribution data, but their performance can degrade significantly in real-world scenarios and when faced with unseen data. In contrast, the proposed MAS offers a universal approach that can be relied upon to a greater extent than fine-tuned models. Furthermore, the fine-tuning process is often demanding, tedious, expensive, and time-consuming, and can be infeasible with small training datasets. The proposed MAS is not encumbered by these challenges, making it a more practical and robust solution.

To address *RQ4* and *RQ5*, we conducted a qualitative analysis of transcripts (samples presented in the appendix) generated by the MAS during the execution of all four experimental tasks. Our findings indicate that agents actively sought consensus, demonstrated opinion modification in response to peer influence, and engaged in critical evaluation - both of their own contributions and those pro-posed by other agents. This behaviour mirrors key characteristics observed in human collaborative debate processes. Furthermore, analysis revealed discernible heterogeneity in the responses, justifications, and argumentation strategies employed by individual agents. We concluded that these observed variations were attributable to the distinct LLMs utilized for each agent.

## 5 Conclusion

The growing threat of fake news and disinformation poses a substantial challenge to contemporary societies and communities. Addressing this complex issue necessitates the development of automated tools capable of both countering false narratives and augmenting the efforts of human fact-checkers. Generative artificial intelligence, particularly LLMs, offers promising avenues for mitigating these challenges; however, realizing their full potential requires dedicated research focused on performance and transparency. This paper presents a novel solution leveraging the MAS built upon open-source LLMs.

An empirical evaluation was conducted through four distinct tasks designed to address five specific research questions pertaining to the performance of this automated tool for tasks related to disinformation detection. The results demonstrate strong performance, achieving a recall score of 0.98 for a medium-resource language and 0.97 for a low-resource language. These findings suggest that the model exhibits comparable capabilities to human annotators in identifying both disinformation and claims worth checking.

Although this system currently does not achieve state-of-the-art performance across all metrics, particularly when compared with fine-tuned models, its evaluation confirms several key strengths. Specifically, the results highlight the system's generalizability, inherent transparency due to its open-source based LLMs, adaptability to novel tasks, and robust performance in small data scenarios. Furthermore, observed performance is not significantly inferior to that of highly optimized models, and even surpasses fine-tuned counterparts on select metrics. This suggests a valuable alternative approach for disinformation detection and analysis.

## Generative AI to support paper writing

We have used the generative AI for grammar, punctuation, vocabulary checks, and to improve the style of the text. We have not used generative AI support in the research process beyond the experiments with MAS and LLMs validation presented in Sections 3 and 4.

## Acknowledgment

This publication is funded by the European Union NextGenerationEU through the Recovery and Resilience Plan for Slovakia under the project No. 09I01-03-V04-00100.